\documentclass[3p,times,procedia]{elsarticle}
\usepackage{ecrc}
\usepackage[bookmarks=false]{hyperref}
    \hypersetup{colorlinks,
      linkcolor=blue,
      citecolor=blue,
      urlcolor=blue}
\usepackage{booktabs}

\volume{00}

\firstpage{1}

\journalname{Procedia Computer Science}

\runauth{Author name}

\jid{procs}

\usepackage{amssymb}

\usepackage[figuresright]{rotating}

\begin{document}
\begin{frontmatter}



\dochead{Proceedings of International Conference on Biomimetic Intelligence and Robotics}

\title{Ethical Decision-making for Autonomous Driving based on LSTM Trajectory Prediction Network}


\author[a,b]{Wen Wei}
\author[a,b,c]{Jiankun Wang\corref{cor1}}

\address[a]{Shenzhen Key Laboratory of Robotics Perception and Intelligence, Shenzhen, China}
\address[b]{Department of Electronic and Electrical Engineering, Southern University of Science and Technology, Shenzhen, China}
\address[c]{Jiaxing Research Institute, Southern University of Science and Technology, Jiaxing, China}

\begin{abstract}
The development of autonomous vehicles has brought a great impact and changes to the transportation industry, offering numerous benefits in terms of safety and efficiency. However, one of the key challenges that autonomous driving faces is how to make ethical decisions in complex situations. To address this issue, in this article, a novel trajectory prediction method is proposed to achieve ethical decision-making  for autonomous driving.  Ethical considerations are integrated into the decision-making process of autonomous vehicles by quantifying the utility principle and incorporating them into mathematical formulas. Furthermore, trajectory prediction is optimized using LSTM network with an attention module, resulting in improved accuracy and reliability in trajectory planning and selection. Through extensive simulation experiments, we demonstrate the effectiveness of the proposed method in making ethical decisions and selecting optimal trajectories. 
\end{abstract}

\begin{keyword}
Autonomous Vehicles; Decision-making; Trajectory Prediction




\end{keyword}
\cortext[cor1]{Corresponding author. }
\end{frontmatter}

\email{wangjk@sustech.edu.cn}



\section{Introduction}
\label{main}
The continuous development of autonomous driving technology has significantly reduced the risk of traffic accidents and greatly improved the efficiency of services such as taxis, buses, and logistics \cite{1,2}. However, considering ethics has become increasingly important in order to ensure higher levels of safety, human-machine interaction, and driving convenience, as they determine how vehicles operate in critical situations. More detailed discussions on ethical dilemmas can be found in the literature on autonomous driving \cite{3,4,5,6}. The government requires the establishment of regulations and safety measures to ensure that autonomous vehicles (AVs) meet specific operational standards for the safety of passengers and other road users \cite{7,8,9}. Therefore, it is essential to optimize traditional planning algorithms without neglecting ethical factors \cite{10,11,12,13}.

\begin{figure}[!t]
\centering
\includegraphics[width=5.5in]{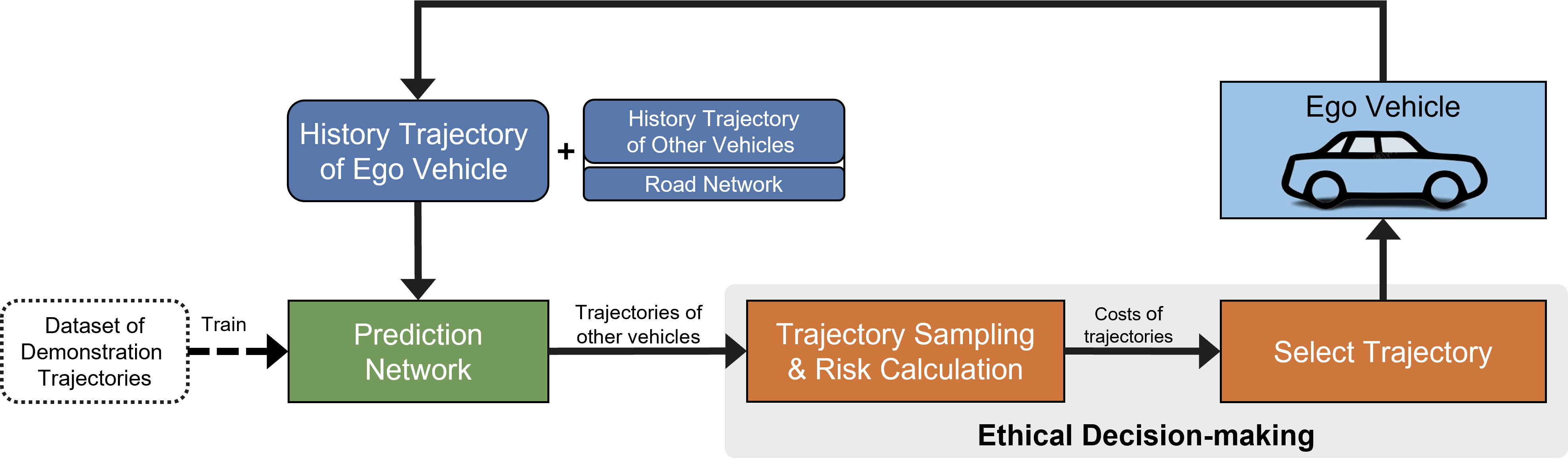}
\caption{Ethical decision-making for autonomous driving based on trajectory prediction network.}
\label{fig4}
\end{figure}

Several studies have explored the ethical aspects of AVs and ethical decision-making in various scenarios. The following works provide valuable insights into the field of AV ethics: Evans et al. \cite{14} present a computational method for evaluating the social acceptability of AV ethical decision-making. Their work considers ethical demands and road user expectations, offering an evaluation tool for decision-makers. Pickering and D’Souza \cite{15} propose a safe and ethical autonomous driving navigation algorithm in highway scenarios based on Deontological Ethics Principles. Awad et al. \cite{16} conduct the Moral Machine experiment, which investigates various ethical dilemmas that AVs would face. They collect over 40 million decisions, providing valuable data for further analysis and ethical considerations. In another work by Evans et al. \cite{17}, they apply Ethical Valence Theory to the decision-making strategy in autonomous driving. This strategy considers parameters such as ethical profiles and the likelihood of injuries, modeling the scenario as a Markov Decision Process. Ebina and Kinjo \cite{18} propose a method based on a social welfare function. Their method considers traffic congestion, ethical decision-making, and stakeholder interests to optimize social costs and maximize social welfare.

\begin{figure}[!t]
\centering
\includegraphics[width=4.5in]{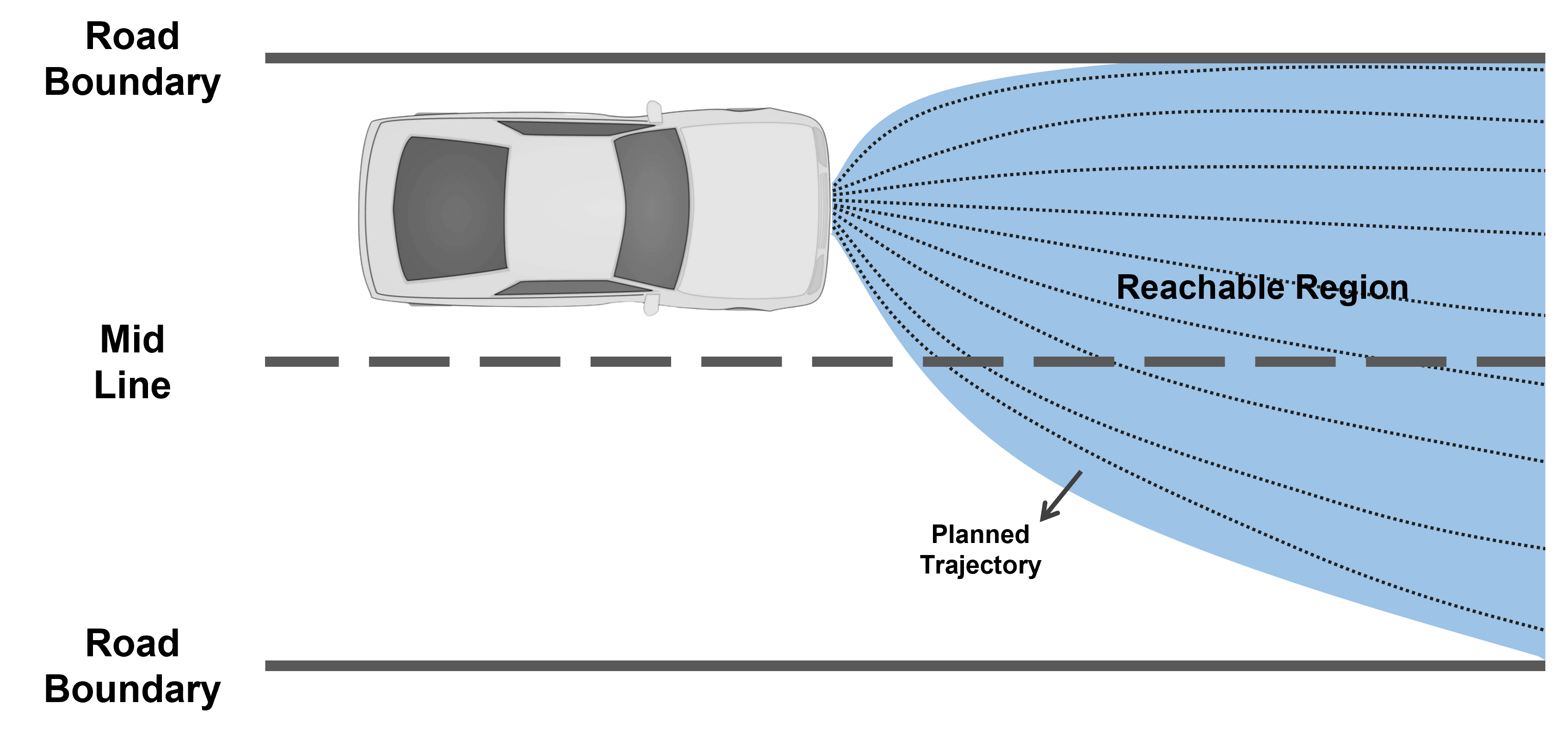}
\caption{Schematic visualization of the trajectory planning. The blue area within the road boundaries represents the reachable region for vehicles, while the dotted lines represent discrete trajectories within that region.}
\label{fig1}
\end{figure}

However, these studies have limitations as they primarily focus on specific ethical dilemmas and decision strategies. In contrast, our work aims to provide a comprehensive approach that includes general trajectory planning and selection for AVs, as shown in Fig. \ref{fig4}. The trajectory prediction model generates multiple trajectories, as shown in Fig. \ref{fig1}, and assigns risk values to each road user based on quantified ethical principles, improving the algorithmic model proposed by Geisslinger et al. \cite{19}. It  selects the trajectory with the lowest execution cost by evaluating and comparing the risks associated with each trajectory.
To enhance the accuracy of trajectory generation, we optimize the underlying trajectory prediction network \cite{20}.  Adding an attention module to the network enhances the accuracy and performance of the trajectory prediction model. The proposed method aims to address various situations while ensuring that AVs' behavior aligns with societal values and ethical principles.

The contribution of this article can be summarized as follows:
\begin{itemize}
\item Integration of ethical considerations: we quantify the influence of ethical factors on AVs behavior and incorporate them into the decision-making process. This is achieved by modeling the utility principle and the ethics of risk as mathematical formulas. 
\item Optimization of the trajectory prediction network with attention module: the trajectory prediction capabilities of the network are enhanced by incorporating an attention module. This attention module considers both trajectory prediction and uncertainty.  The network is empowered to effectively weigh and select potential trajectories while considering the associated level of uncertainty.  This optimization improves the accuracy and reliability of trajectory planning and selection for AVs.

\end{itemize}    

The rest of this article is organized as follows. Section II quantifies the influence of ethical factors. Section III describes the optimized attention-based network model. Section IV and V presents the results of the simulation experiments and concludes this work, respectively. 

\section{Integration of Utility Principle}
\subsection{Ethics of Risk}

Autonomous driving technologies operate in complex and highly dynamic road conditions, requiring instantaneous decisions that may have a significant impact on the safety and well-being of passengers, and other road users. Ethics of risk provides a framework in this context for assessing ethical issues from uncertainty and potential risks. In the ethics of risk, harm is defined as a numerical value to measure the impact on different groups when the collision happens, while risk is the product of collision probability and harm. By employing the principle of quantified utility, we calculate the risk values of different trajectories and select the trajectory with the minimum risk.

\subsection{Utility Principle}

In the case of autonomous driving, the utility principle aims to make decisions that maximize overall welfare. For example, in emergency steering situations, self-driving cars follow the utility principle, prioritizing overall safety while minimizing harm to passengers, and other road users like pedestrians and vehicles. When pedestrian crossings are involved, self-driving cars follow the utility principle, prioritizing the safety and well-being of pedestrians.

In response to potential hazards or risks, we design a utility cost function that weighs the quality between different trajectory options. The function takes into account factors such as the likelihood and severity of injury, as well as the utility principle. The total utility cost is represented by the following  function:
\begin{equation}
\label{equation2}
\displaystyle \rm J_{\mathrm{utility}} =\left\{
\begin{array} {cl}
0 ,& \mathrm{  J(s) \leq J_{mean}(s)}\\  
\mathrm{J}(s) ,& \mathrm{otherwise} \\
\end{array} \right.
\end{equation}

\begin{equation}
\label{equation3}
\displaystyle \rm { J_{mean}}(s) = \frac{{\sum_{i =1}^{ N_r} R_i(s) + \sum_{j =1}^{ N_j} R_j(s)}}{{\left| N_r\right|}} 
\end{equation}

\begin{equation}
\label{equation4}
\displaystyle  \rm J(s) = {{\sum_{i{=1}}^{ S_r} R_i(s) + \sum_{j {=1}}^{ S_j} R_j(s)}}
\end{equation}

\begin{equation}
\label{equation1}
    \rm J_{risk}(s)=\omega_{o}J_{origin}+\omega_{u}J_{utility}
\end{equation} 

This formula involves two components: {$\rm J_{mean}(s)$} and {$\rm J(s)$}. {$\rm J_{mean}(s)$} is a reference value that represents the maximum utility cost limit under given conditions, which is described by equation (\ref{equation3}). In these equations, R represents the risk value associated with this trajectory while s represents the trajectory. It weighs the individual contributions of ego risk and obstacle risk based on the number of elements in the ego risk vector ({$\rm N_r$} for ego risk) and divides the result by the sum of the absolute values of ego risk. {$\rm J(s)$}, as shown in equation (\ref{equation4}), represents the actual utility cost in the current environment, calculated by weighting and summing ego risk and obstacle risk. The decision-making process then compares {$\rm J(s)$} and {$\rm J_{mean}(s)$} to make decisions, as shown in equation (\ref{equation2}). If {$\rm J(s)$} does not exceed {$\rm J_{mean}(s)$}, it indicates that the current decision satisfies the utility principle cost, and no further action is required. However, if {$\rm J(s)$} exceeds {$\rm J_{mean}(s)$}, it suggests that the current decision has issues and additional measures are required to reduce the utility cost. The function returns the result of {$\rm J_{mean}(s)$} in this situation.

By applying the utility principle cost, AVs can make decisions by weighing and comparing different trajectory options based on their associated risks. The objective is to minimize the overall utility cost while balancing safety and minimizing potential harm to pedestrians and other road users. Optimizing decisions allows the autonomous driving system to adhere to the utility principle cost and make decisions that maximize overall benefits in various scenarios.

Through the design of a utility principle cost function, the original cost function is augmented to incorporate it as a measure of the trajectory. The trajectory is represented by equation (\ref{equation1}), where the weighting factor {$\omega$} determines the proportion of the corresponding principle.
To further refine the decision-making process, we assign risk values to each road user and each trajectory based on quantified ethical principles. This involves considering factors such as the probability of injury and the severity of potential harm caused by each trajectory. By evaluating and comparing the risk values associated with each trajectory, we can determine the trajectory that minimizes the overall cost.

In summary, the utility principle cost plays a crucial role in autonomous driving as it guides the ethical evaluation system and decision-making process, which results in optimizing the decision-making process. This comprehensive consideration ensures that AVs adhere to ethical principles while prioritizing the interests and safety of all parties involved.

\section{Attention Based LSTM Trajectory Prediction}
As mentioned above, our ethical quantification is based on the concepts of risk and harm. In order to maximize the effectiveness of the quantified utility principle formula, it is essential to accurately calculate collision probabilities and optimize trajectory predictions.

Adding an attention module on top of the original Long Short Term Memory (LSTM) networks allows the model to better focus on relevant historical trajectory information, thereby improving the accuracy and performance of the predictions. In vehicle trajectory prediction tasks, historical trajectories consist of a large amount of time-series data. However, not all historical trajectory points are equally important for predicting future trajectories. Some points have higher relevance due to their proximity to the prediction time or their significance in certain situations. The attention mechanism enhances the encoding of historical trajectories by learning the weights of different trajectory points, which is achieved by incorporating it into the LSTM network. At each time step, the LSTM network generates an attention vector based on the input and hidden state. This attention vector is then used to weigh the hidden states of the historical trajectories. Let {$\rm H$} be the hidden state of the historical trajectory of the ego vehicle, and {$\rm N = (H_1, H_2, ..., H_{ m})$} be the hidden states of the historical trajectories of other obstacles, where {$\rm m$} represents the number of obstacles. To calculate the attention vector {$\rm A_{ i}$} for each historical trajectory point {$\rm H_{ i}$}, we can use the cosine similarity \cite{20} between the current input and hidden state vectors {$\rm X$} and the hidden state {$\rm H_{ i}$}:

 \begin{equation}
    \begin{array}{lcl}
    \displaystyle  \rm A_{ i} = \frac{\rm \cos(X, H_{ i})}{\sum_{ {j}\rm{=1}}^{\rm m}\cos(X, H_{ j})}
    \end{array}
\end{equation}

By multiplying the attention weights with the historical trajectory hidden states and summing them, a weighted representation of the historical trajectories is obtained. This allows the model to focus more on the relevant historical trajectory points.
\begin{equation}
\label{6}
    \begin{array}{lcl}
    \displaystyle  \rm W_H = {\sum_{ i \rm{=1}}^{\rm m} (A_{ i} \cdot \rm H_{ i})}
    \end{array}
\end{equation}

\begin{figure}[!t]
\centering
\includegraphics[width=5.5in]{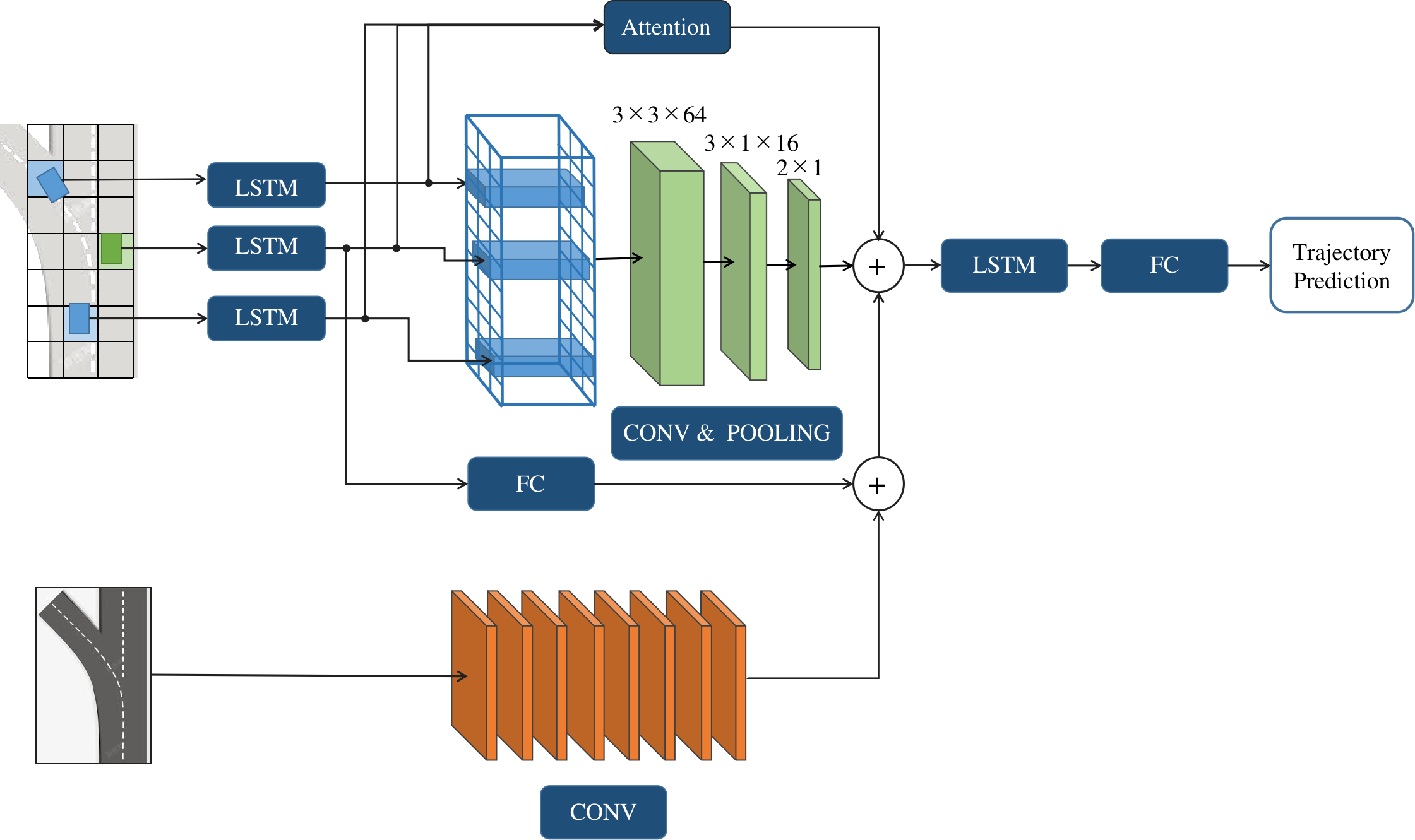}
\caption{Optimized attention-based LSTM network for the trajectory prediction model.}
\label{fig2}
\end{figure}

The obtained result  {$\rm W_{ H}$} in equation (\ref{6}) is fused with the social convolution pool layer, and the results are as follows:
\begin{equation}
    \displaystyle \rm E = \mbox{cs\_pool}\rm (H_1, H_2, ..., H_{ m}) + W_{ H}
\end{equation}
where \mbox{cs\_pool} is the social convolutional pooling network. To continue, we input E into the decoder. Finally, the optimized attention-based LSTM network for the trajectory prediction model is shown in Fig. \ref{fig2}.

While the original network structure can partially consider the importance of historical trajectory points, the attention mechanism further enhances this ability. It avoids treating all historical trajectory points equally by adaptively learning their importance weights based on different input features and contexts. This allows the model to incorporate important historical trajectory points more accurately into the prediction process.

\section{Results of Simulation Experiments}
\begin{figure}[!t]
\centering
\includegraphics[width=6 in]{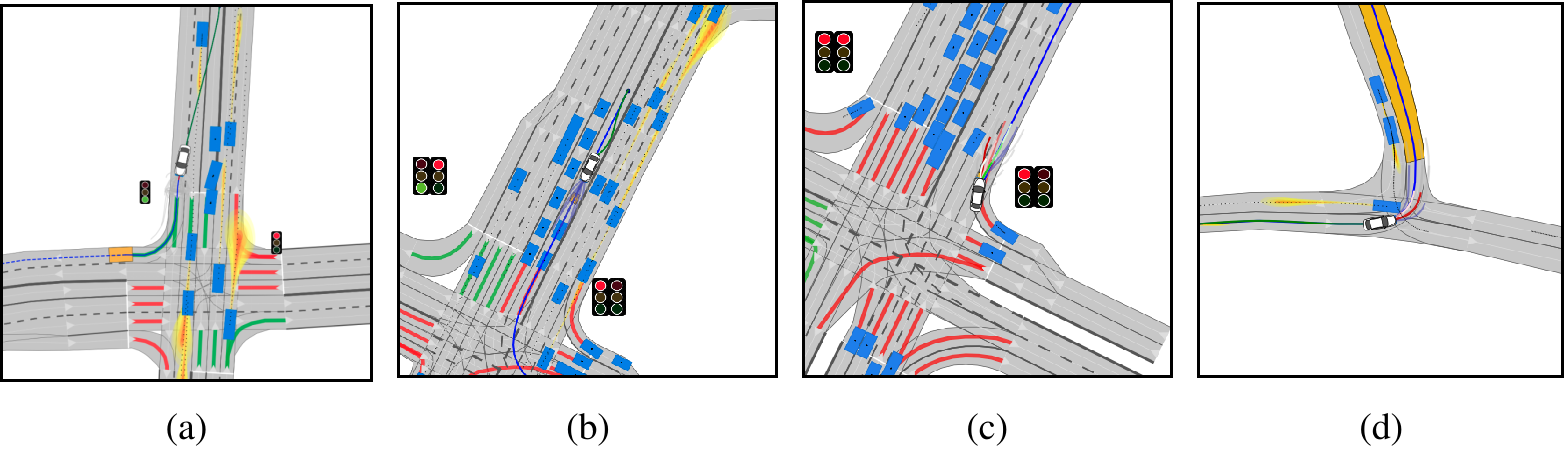}
\caption{Examples of CommonRoad scenarios. In these scenarios, blue squares represent other vehicles, while yellow squares represent the destination. The four scenarios shown are (a) vehicle lane changing, (b) encountering traffic congestion, (c) right-turning, and (d) driving in the opposite direction, which are complex situations that autonomous driving faces. }
\label{fig3}
\end{figure}

In this section, we conduct simulation experiments using the CommonRoad dataset \cite{21} to evaluate the performance of the proposed method in ethical decision-making and trajectory selection. To ensure comparability, we adopt the same planning methods and run these two algorithms in 2000 simulated scenarios from the CommonRoad dataset. These scenarios cover road networks and traffic situations from multiple countries, as shown in Fig. \ref{fig3}, including Germany, the United States, China, and Japan. The scenes are either based on real-world recordings or manually created, encompassing various driving environments such as highways, urban roads, and rural roads to simulate different traffic conditions.

By comparing the risks and simulated accidents brought by different planning methods, we focus primarily on three categories of road users: the Ego-AV, all other road users excluding the autonomous vehicle (Third Party), and vulnerable road users (VRUs). We quantify the harm value resulting from each collision and calculate the total harm and calculate the harm distribution based on the different groups, as shown in Table \ref{table1}. The results indicate that explicitly considering risks in trajectory planning can reduce risks for all road users. Compared with the original method, ours performs better in ethical decision-making in terms of completion rate and total harm.   The completion rate increases from 70.70\% to 77.95\%, indicating more successful completion of tasks in the simulated scenarios. At the same time, our method can reduce the total harm value from 22.35 to 18.16, demonstrating significant improvement in the safety of road users. 
These results can be attributed to the consideration of both ego and surrounding vehicle risks in the utility principle and the improved attention mechanism in the network model. By considering the overall risk and actively attending to the information in the surrounding environment, our method leads to better decision-making and more appropriate trajectory selection in dynamic scenarios.

\begin{table}[h]
\centering
\begin{tabular}{ccccccc}
\toprule
Algorithm & Completed rate ({\it{\%}}) & Total harm & Ego-AV & Third party & VRUs\\
\midrule
Origin  &   70.70 &  22.35&   11.50 &  10.84 & 3.11\\
\textbf{Ours}  & 77.95 &  18.16& 9.86 &  8.30 & 3.45\\
\bottomrule
\end{tabular}
\caption{\centering Performance Comparison Table}
\label{table1}
\end{table}

The distribution of harm is an important aspect that reflects the ethical nature of the method. Our goal is to minimize harm distribution among road users while simultaneously safeguarding the interests of vulnerable groups.
In terms of harm distribution, our method performs better in reducing harm values, as shown in Table \ref{table1}.  Compared with the original method, our method reduces the harm value caused by the autonomous vehicle from 11.50 to 9.86, and the harm value caused by the third party is reduced from 10.84 to 8.30. This indicates that the proposed method makes significant progress in reducing risks associated with AVs as well as risks caused by other road users. The proposed method slightly increased the harm value caused by VRUs from 3.11 in the original method to 3.45, indicating that our method maintains relative stability in harm values caused by VRUs.  The change could be attributed to the utility principle's focus on the overall harm distribution and higher weight ($\rm \omega_u $) assigned in the trajectory selection formula. It is possible that the current weighting of $\rm \omega_u $ might prioritize considerations that result in higher harm for VRUs.

The experimental results demonstrate the effectiveness of the proposed method. In terms of ethical decision-making, our method consistently outperforms the baseline method. Not only does it increase the success rate, but it also reduces the distribution of harm values generated. This indicates that incorporating the utility principle into the decision-making process improves the overall ethical behavior of AVs. In terms of trajectory selection, the optimized LSTM network with an attention module shows significant improvements. The enhanced trajectory selection contributes to improving trajectory planning and the overall safety of AVs. To achieve a more balanced and universal ethical framework, it would be necessary to adjust the weight ($\rm \omega_u $) and fine-tune the utility principle accordingly. By carefully calibrating the weights assigned to different factors in the decision-making process, we can strike a better balance between the interests of different road users, including VRUs.

\section{Conclusions}
In conclusion, we present a novel method that combines ethical considerations with optimized trajectory prediction for AVs. The experimental results demonstrate the effectiveness of the proposed method in enhancing ethical decision-making and improving overall safety. By integrating ethical factors into the decision-making process and optimizing trajectory selection using an LSTM network with an attention module, we achieve better performance in terms of success rate and harm reduction. In the near future, we plan to conduct real-world experiments to further evaluate the proposed method and investigate other factors that contribute to ethical decision-making for autonomous driving.

\section*{Acknowledgements}

This work is supported by the Shenzhen Key Laboratory of Robotics Perception and Intelligence under Grant ZDSYS20200810171800001, Shenzhen Outstanding Scientific and Technological Innovation Talents Training Project under Grant RCBS20221008093305007, and National Natural Science Foundation of China grant \# 62103181.

\clearpage

\normalMode

\end{document}